\documentclass[journal]{myIEEEtran}
\usepackage{cite}
\usepackage{amsmath,amssymb,amsfonts}
\usepackage{algorithmic}
\usepackage{graphicx,color}
\usepackage{textcomp}
\usepackage{hyperref}
\usepackage{balance}
\def\BibTeX{{\rm B\kern-.05em{\sc i\kern-.025em b}\kern-.08em
    T\kern-.1667em\lower.7ex\hbox{E}\kern-.125emX}}
\AtBeginDocument{\definecolor{ojcolor}{cmyk}{0.93,0.59,0.15,0.02}}

\usepackage{pifont}
\usepackage{booktabs}
\usepackage{arydshln}
\usepackage{url}
\usepackage{xspace}
\usepackage{soul}
\usepackage{fontawesome5} % only used for GitHub icon, can be removed

\newcommand{\cmark}{\ding{51}}%
\newcommand{\xmark}{\ding{55}}%
\newcommand{\ie}{\emph{i.e.,}\xspace}
\newcommand{\eg}{\emph{e.g.,}\xspace}

\newcommand{\seg}{segmentation\xspace}
\newcommand{\segs}{segmentations\xspace}
\newcommand{\anntr}{annotator\xspace}
\newcommand{\anntrs}{annotators\xspace}
\newcommand{\anntn}{annotation\xspace}
\newcommand{\anntns}{annotations\xspace}

\newcommand{\intanntagr}{inter-annotator agreement\xspace}

\newcommand{\newdatasetnamefull}{ISIC MultiAnnot++\xspace}
\newcommand{\newdatasetname}{IMA++\xspace}
\newcommand{\isicar}{ISIC Archive\xspace}
\newcommand{\numimgs}{14,967\xspace}
\newcommand{\numimgswithoutmetadata}{59\xspace} % -> 15026-14967
\newcommand{\numanntrs}{16\xspace}
\newcommand{\numimgswithsinglesegs}{12,573\xspace}
\newcommand{\numimgswithmultisegs}{2,394\xspace}
\newcommand{\numsegstotal}{17,684\xspace} % -> after sanitization
\newcommand{\numsegstotalwconsensus}{22,472\xspace} % -> with consensus (ST, MV)
\newcommand{\numallblacksegs}{59\xspace} % -> "Empty mask"
\newcommand{\numallwhitesegs}{3\xspace} % -> "Mask covers entire image"
\newcommand{\numsegstouchingborder}{1,129\xspace} % -> "Mask touches image border"
\newcommand{\githubrepo}{\url{https://github.com/sfu-mial/IMAplusplus}}

\newcommand{\say}[1]{``#1''}
\title{\newdatasetname: ISIC Archive Multi-Annotator Dermoscopic Skin Lesion Segmentation Dataset}

\author{Kumar Abhishek$^\dagger$, Jeremy Kawahara$^\ddagger$, and Ghassan Hamarneh$^\dagger$\thanks{Corresponding author: kabhishe@sfu.ca (Kumar Abhishek)}
\\
$^\dagger$Medical Image Analysis Lab, School of Computing Science, 
Simon Fraser University, Canada
\\
$^\ddagger$AIP Labs, Hungary}

\begin{document}

\maketitle

%\receiveddate{00 April, 2024}
%\reviseddate{11 April, 2024}
%\accepteddate{00 April, 2024}
%\publisheddate{00 May, 2024}
%\currentdate{24 June, 2024}
%\doiinfo{DD.2024.0916000}

% \author{FIRST A. AUTHOR\authorrefmark{1} (FELLOW, IEEE), SECOND B.
% AUTHOR\authorrefmark{2}, AND THIRD C. AUTHOR,~JR.\authorrefmark{1,2}
% (MEMBER, IEEE)}
% \affil{National Institute of Standards and 
% Technology, Boulder, CO 80305 USA}
% \affil{Department of Physics, Colorado State University, Fort Collins, 
% CO 80523 USA}
% \authornote{The authors contributed equally to this article. This work was supported by the XYZ Research Council.}
% \markboth{DESCRIPTOR: ISIC ARCHIVE MULTI-ANNOTATOR DERMOSCOPIC SKIN LESION SEGMENTATION DATASET (\newdatasetname) }{ABHISHEK ET AL.}

\begin{abstract}
Multi-\anntr medical image segmentation is an important research problem, but requires annotated datasets that are expensive to collect.
Dermoscopic skin lesion imaging allows human experts and AI systems to observe morphological structures otherwise not discernable from regular clinical photographs.
However, currently there are no large-scale publicly available multi-\anntr skin lesion segmentation (SLS) datasets with annotator-labels for dermoscopic skin lesion imaging.
We introduce \newdatasetnamefull, a large public multi-\anntr skin lesion segmentation dataset for images from the ISIC Archive. 
%The final dataset contains \numimgs images with one or more segmentations per image. 
The final dataset contains \numsegstotal segmentation masks spanning \numimgs dermoscopic images, where
%Of these images, 
\numimgswithmultisegs dermoscopic images have 2-5 \segs per image, %resulting in a total of \numsegstotal \seg masks, 
making it the largest publicly available SLS dataset. 
Further, metadata about the segmentation, including the annotators' skill level and segmentation tool, is included, enabling research on topics such as    
%The \segs contain metadata corresponding to the annotators' skill levels as well as the tool used to perform the \seg, enabling several kinds of research including, but not limited to, 
\anntr-specific preference modeling for segmentation and \anntr metadata analysis.
We provide an analysis on the characteristics of this dataset, curated data partitions, and consensus segmentation masks.
% \\ 
%  \\ 
%  {\textcolor{ieeedata}{\abstractheadfont\bfseries{IEEE SOCIETY/COUNCIL}}}     Engineering Medicine \& Biology Society (EMBS)\\  
%  \\
%  {\textcolor{ieeedata}{\abstractheadfont\bfseries{DATA DOI/PID}}}     \href{https://doi.org/10.5281/zenodo.14201692}{10.5281/zenodo.14201692}\\ 
  
%  {\textcolor{ieeedata}{\abstractheadfont\bfseries{DATA TYPE/LOCATION}}}  segmentation masks; medical image segmentation; 
% % Lesion, Melanoma, Annotator agreement, 
\end{abstract}

\begin{IEEESociety}
    Engineering Medicine \& Biology Society (EMBS)
\end{IEEESociety}

\vspace{0.75ex}

\begin{DataDOI}
    \faDatabase \href{https://doi.org/10.5281/zenodo.14201692}{10.5281/zenodo.14201692}
\end{DataDOI}

\vspace{0.75ex}

\begin{CodeRepo}
    \faGithub \href{https://github.com/sfu-mial/IMAplusplus}{sfu-mial/IMAplusplus}
\end{CodeRepo}

\vspace{0.75ex}

\begin{DataType}
    \seg masks; medical image \seg
\end{DataType}

\vspace{0.75ex}


\IEEEoverridecommandlockouts
\IEEEpubid{\begin{minipage}[t]{\textwidth}\ \\[10pt]
        \centering\footnotesize{\copyright 2025. This manuscript version is made available under \href{http://arxiv.org/licenses/nonexclusive-distrib/1.0/}{the arXiv.org perpetual, non-exclusive license 1.0}.}
\end{minipage}}

\section{BACKGROUND}

Skin cancer is the most common form of cancer diagnosed globally, with an estimated 1.5 million new diagnoses in 2022 alone, according to the World Health Organization~\cite{whoiarc}. Melanoma, the most aggressive form of skin cancer, alone accounted for 330,000 new diagnoses and 60,000 deaths that same year. With the incidence rate of skin cancers rising driven by factors including but not limited to: ageing populations, rising life expectancy, increased sun exposure, and environmental factors~\cite{wang2025burden,zhou2025gloabl}, the incidence rate of melanoma is projected to double every 10-20 years~\cite{leiter2014epidemiology}. This burgeoning global burden of skin diseases, estimated to be over 42 million DALYs (disability-adjusted life years) in 2019~\cite{yakupu2023burden}, coupled with projections of a declining dermatologist-to-population ratio~\cite{lim2017burden}, strongly motivates the research and development of automated methods for dermatological image analysis.

\begin{table}[!htbp]
\centering
\caption{Comparing public single- and multi-annotator skin lesion image \seg datasets. \newdatasetname spans the largest number of images and has the largest number of \segs.}
\label{tab:datasets-comparison}
\resizebox{\linewidth}{!}{%
\def\arraystretch{1.25}
\begin{tabular}{@{}cccccc@{}}
\toprule
\textbf{Dataset}       & \textbf{Year} & \textbf{Modality}   & \begin{tabular}[c]{@{}c@{}}\textbf{Total Images}\\(train/val/test)\end{tabular} & \begin{tabular}[c]{@{}c@{}}\textbf{Multi-Annotator}\\\textbf{Segmentations?}\end{tabular} & \begin{tabular}[c]{@{}c@{}}\textbf{Total}\\\textbf{Segmentations}\end{tabular} \\ \midrule
SCD~\cite{Skin_Cancer_Detection_data}           & 2013 & Clinical   & \begin{tabular}[c]{@{}c@{}}206\\ (N/A)\end{tabular}                     & \xmark                                                                  & 206                                                           \\
\hdashline
DermoFit~\cite{ballerini2013color}      & 2013 & Clinical   & \begin{tabular}[c]{@{}c@{}}1,300\\ (N/A)\end{tabular}                   & \xmark                                                                  & 1,300                                                         \\
\hdashline
PH2~\cite{mendonca2013ph}           & 2013 & Dermoscopy & \begin{tabular}[c]{@{}c@{}}200\\ (N/A)\end{tabular}                     & \xmark                                                                  & 200                                                           \\
\hdashline
ISIC 2016~\cite{gutman2016skin}     & 2016 & Dermoscopy & \begin{tabular}[c]{@{}c@{}}1,279\\ (900/-/379)\end{tabular}             & \xmark                                                                  & 1,279                                                         \\
\hdashline
ISIC 2017~\cite{codella2018skin}     & 2017 & Dermoscopy & \begin{tabular}[c]{@{}c@{}}2,750\\ (2,000/150/600)\end{tabular}         & \xmark                                                                  & 2,750                                                         \\
\hdashline
ISIC 2018~\cite{codella2019skin}     & 2018 & Dermoscopy & \begin{tabular}[c]{@{}c@{}}3,694\\ (2,594/100/1000)\end{tabular}        & \xmark                                                                  & 3,694                                                         \\
\hdashline
HAM10000~\cite{tschandl2020human,ham10ksegmentations}      & 2020 & Dermoscopy & \begin{tabular}[c]{@{}c@{}}10,015\\ (N/A)\end{tabular}                  & \xmark                                                                  & 10,015                                                        \\
\hdashline
ISIC 2019-Seg~\cite{zepf2023label} & 2023 & Dermoscopy & \begin{tabular}[c]{@{}c@{}}100\\ (N/A)\end{tabular}                     & \cmark                                                                  & 300                                                           \\
\hdashline
\textbf{\newdatasetname}         & 2025 & Dermoscopy & \begin{tabular}[c]{@{}c@{}}\numimgs\\ (X/Y/Z)\end{tabular}                & \cmark                                                                  & \numsegstotal                                                         \\ \bottomrule
\end{tabular}%
}
\end{table}

Dermoscopy is a widely-used non-invasive imaging technique for the examination of pigmented skin lesions, allowing clinicians to visualize both morphological surface features and subsurface structures otherwise obscured to the naked eye~\cite{binder1995epiluminescence,argenyi1997dermoscopy,kittler2002diagnostic}. Studies have shown that when used by trained experts, dermoscopy significantly improves both the sensitivity~\cite{vestergaard2008dermoscopy,rosendahl2012impact} and specificity~\cite{carli2004improvement,vanderrhee2011impact} of melanoma diagnosis. Dermoscopy has also been the target modality for automated skin image analysis. For example, almost 3 decades ago, Binder et al.~\cite{binder1994application} used  artificial neural networks for detecting malignant melanomas from dermoscopic skin lesion images.
% We direct the interested readers to the survey by Korotkov and Garcia~\cite{korotkov2012computerized} for a comprehensive review of automated skin image analysis.

Segmentation of skin lesions is a crucial task in the automated skin lesion image analysis pipeline. Rule-based diagnosis clinical prediction rules, including the most widely used~\cite{forsea2017impact,harrington2017diagnosing} ABCD (Asymmetry, Border, Color, Differential structure)~\cite{nachbar1994abcd}, rely on an accurate delineation of the skin lesion boundary. In recent years, deep learning (DL)-based skin image analysis methods rely on \seg either as an end-goal, an intermediate task (\eg analyzing wide-field images, tracking the evolution of skin lesions, removing imaging artifacts, and enhancing the interpretability of DL models), or as a benchmark for evaluating massive foundation models~\cite{yan2025multimodal,xu2025dermino}. We direct the interested reader to comprehensive surveys on automated skin image analysis in general~\cite{korotkov2012computerized} and DL-based skin lesion segmentation (SLS) in particular~\cite{mirikharaji2023survey}. However, although crucial, SLS remains a challenging task due to the presence of imaging artifacts (\eg hair, gel bubbles, dark corners), lesion size and shape variability, varying skin tones, variable contrast and illumination, and ambiguous lesion boundaries~\cite{mirikharaji2023survey}, all of which affect the annotation of a true ground truth \seg. This variability in medical image \seg, including in skin lesion images, is an active area of research, spanning several related yet distinct goals: studying variability in expert \segs~\cite{sampat2006measuring,silletti2009variability,li2010estimating,fortina2012where,ribeiro2019handling}, aggregating multiple \segs to model a single \say{gold standard} \seg~\cite{warfield2004simultaneous,kats2019soft,ribeiro2020less,mirikharaji2021d,amit2023annotator,hamzaoui2023morphologically}, learning to model individual annotator-specific \seg preferences~\cite{zepf2023label,ji2021learning,liao2024modeling}, modeling the underlying distribution of \segs~\cite{kohl2018probabilistic,baumgartner2019phiseg,rahman2023ambiguous,schmidt2023probabilistic} and discovering the underlying \seg styles~\cite{abhishek2025segmentation}, among others.

Table~\ref{tab:datasets-comparison} lists all the publicly available SLS datasets, spanning the two popular skin imaging modalities: dermoscopic and clinical. Existing datasets' sizes vary from a little as 200 images (PH2~\cite{mendonca2013ph}) to just over 10,000 images (HAM100000~\cite{ham10ksegmentations}), but despite the inherit ambiguity in \seg, all these datasets contain only one \seg mask per image. There is only one SLS dataset that contains multiple \anntns per image: ISIC 2019-Seg~\cite{zepf2023label}, but with only 3 \segs for each of its 100 images, it is quite small for effectively modeling annotator-specific tasks, especially compared to multi-\anntr datasets from other medical imaging modalities such as CT (\eg LIDC-IDRI~\cite{armato2011lung}) and fundus photography (\eg RIGA~\cite{almazroa2017agreement}).

The \isicar, maintained by the International Skin Imaging Collaboration (ISIC), hosts the world's largest collection of digital skin images. At the time of writing this article, the \isicar contains over 1.2 million images, of which more than 120,000 are publicly available dermoscopic images. The Archive also contains subsets that were released as part of ISIC's \say{Skin Lesion Analysis Towards Melanoma Detection Segmentation Challenges} over the years (2016--2018). In fact, a recent survey by Mirikharaji et al.~\cite{mirikharaji2023survey} found that of the 177 papers on SLS that they reviewed, 168 ($\sim$95\%) of the papers used at least one ISIC dataset, underlining the importance of the \isicar to the skin image analysis community.

Motivated by the lack of large multi-\anntr SLS datasets and the popularity of \isicar, we collect and publicly release \newdatasetnamefull (\textbf{\newdatasetname} hereafter). With \numimgs images segmented by \numanntrs annotators, the \newdatasetname contains a total of \numsegstotal \segs. Of these \numimgs images, \numimgswithmultisegs have at least 2 \segs per image.
To the best of our knowledge, \newdatasetname is the largest publicly available skin lesion segmentation dataset, multi-\anntr or otherwise. Additionally, to establish \seg consensus for images that have multiple \segs, we also include \seg masks using two consensus algorithms, increasing the number of \segs to \numsegstotalwconsensus.
The \newdatasetname dataset presents the following meritorious properties:

\begin{figure}[!ht]
\centerline{\includegraphics[width=0.98\linewidth]{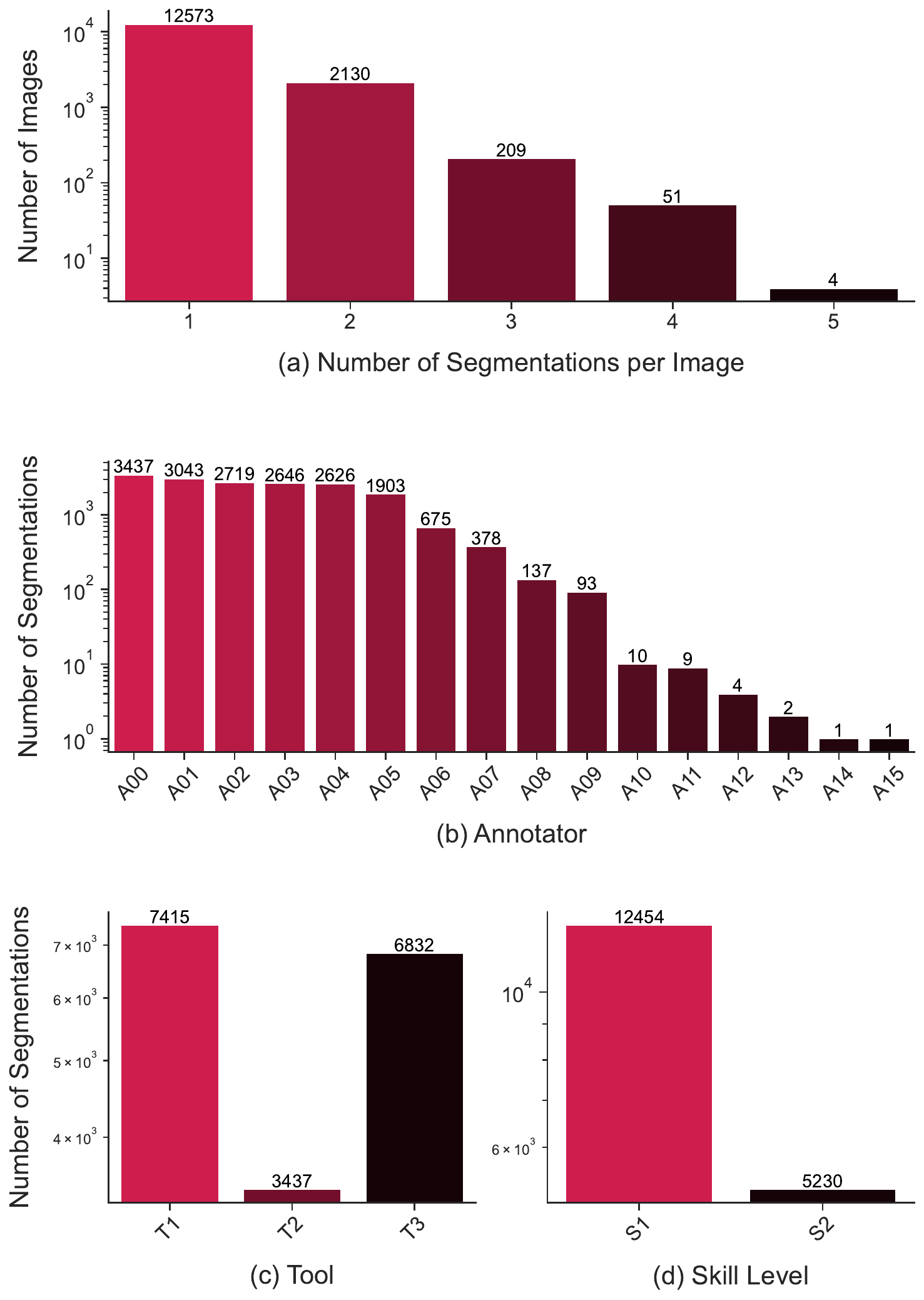}}
\caption{
A breakdown of the \newdatasetname dataset: (a) distribution of number of \segs per image and annotation factor-wise \seg counts: (b) annotator, (c) tool, and (d) skill level.
\label{fig:image-seg-counts}
}
\end{figure}

\begin{itemize}
    \item \textbf{Inter-annotator variability:} \newdatasetname captures a wide range of segmentation styles, reflecting differences in \anntr preferences, tools used, and manual review process (varying skill levels).
    \item \textbf{Realistic multi-\anntr scenario:} Most multi-\anntr medical image \seg datasets~\cite{armato2011lung,fraz2012ensemble,carass2017longitudinal,commowick2018objective,li2024qubiq,carmo2025manual} have every image segmented by every annotator, thus forming a complete bipartite graph between the set of images and the set of annotators. \newdatasetname, on the other hand, simulates real-world \anntn scenarios where multiple \anntrs contribute to a subset of images, and therefore features an incomplete bipartite graph. This means that every image in \newdatasetname is segmented by at least one \anntr, but not all images are segmented by all \anntrs.
    \item \textbf{Tool-specific \seg styles:} Because of the availability of tool and skill level information for the \segs, \newdatasetname allows for the exploration of how variations in these often-overlooked factors affect \seg variability.
\end{itemize}
% We believe that \newdatasetname's large scale combined with expert-provided annotations makes it
These features make \newdatasetname 
an extremely valuable dataset for researchers working on a variety of open problems including: (1) skin lesion image classification; (2) skin lesion \seg, \eg multi-\anntr \seg preference modeling, multi-expert \seg consensus modeling, learning the distribution of \segs and discovering the underlying \seg styles from multi-\anntr masks, and studying inter-\anntr agreement among experts; and, (3) multi-modal (dermoscopic images and rich metadata) and multi-task (diagnosis, segmentation, IAA prediction) skin image analysis, and other tasks.
% \begin{itemize}
%     \item skin lesion image classification,
%     \item skin lesion \seg:
%     \begin{itemize}
%         \item multi-\anntr \seg preference modeling,
%         \item multi-expert \seg consensus modeling,
%         \item learning the distribution of \segs and discovering the underlying \seg styles from multi-\anntr masks,
%         \item studying inter-\anntr agreement among experts, and
%     \end{itemize}
%     \item multi-modal (dermoscopic images and rich metadata) and multi-task (diagnosis, segmentation, IAA prediction) skin image analysis, and other tasks.
% \end{itemize}

Subsets of \newdatasetname have been used in two recent papers: \textbf{(a)} to evaluate the discovery of unique \anntn \seg styles in the absence of \anntr-\seg correspondence~\cite{abhishek2025segmentation} and \textbf{(b)} to examine a statistical association between the \textbf{\underline{i}}nter-\textbf{\underline{a}}nnotator \seg \textbf{\underline{a}}greement levels (IAA hereafter) and the malignancy of skin lesions and to evaluate the feasibility of predicting IAA from skin lesion images directly without requiring any \segs~\cite{abhishek2025what}. 

\begin{figure*}[!htbp]
\centerline{\includegraphics[width=0.95\linewidth]{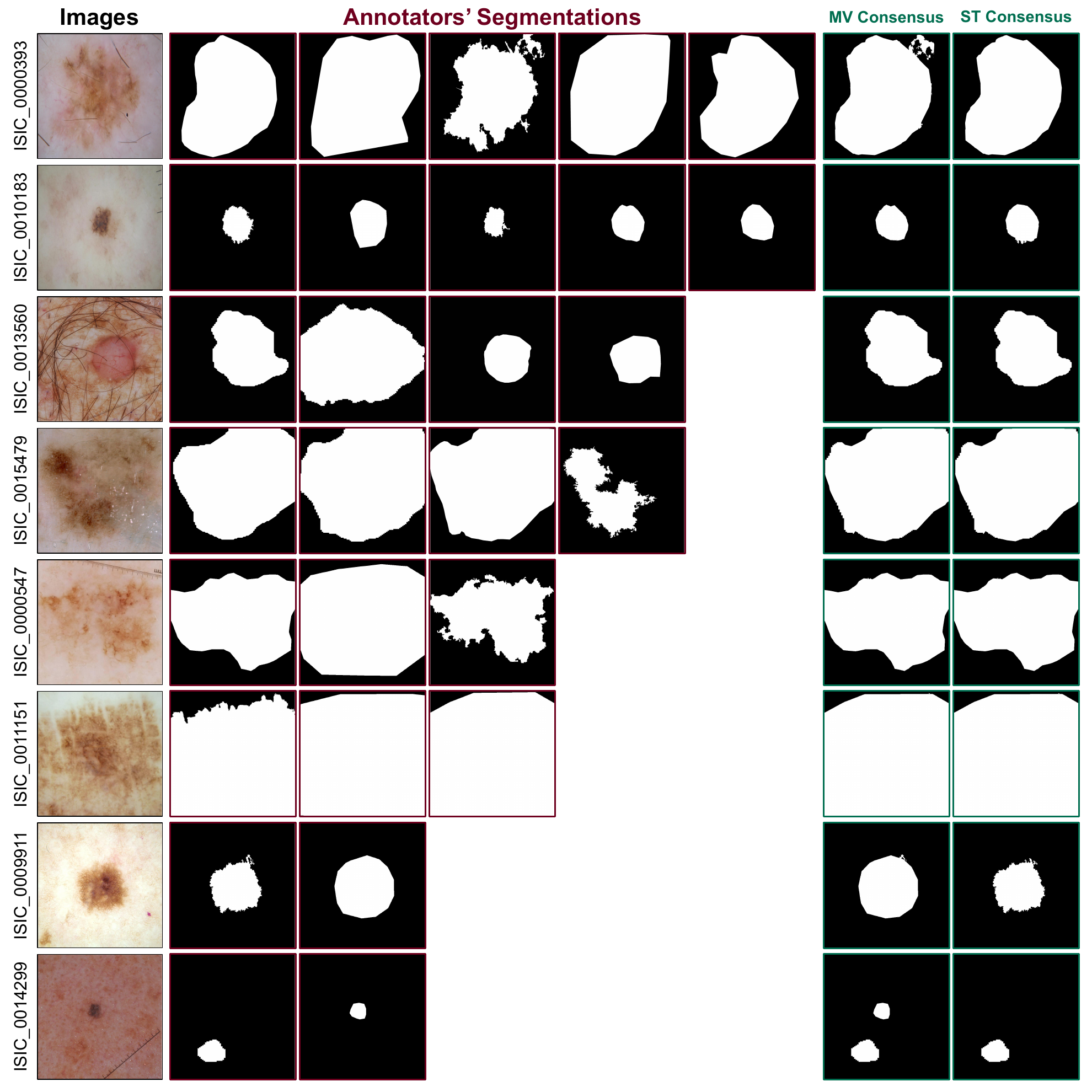}}
\caption{
Sample image-\seg pairs from \newdatasetname: 2 rows each for images with \{5, 4, 3, 2\} \segs per image along with the corresponding consensus \seg masks computed using majority voting (MV) and STAPLE (ST).
\label{fig:dataset-samples}
}
\end{figure*}

\begin{figure*}[!htbp]
\centerline{\includegraphics[width=\linewidth]{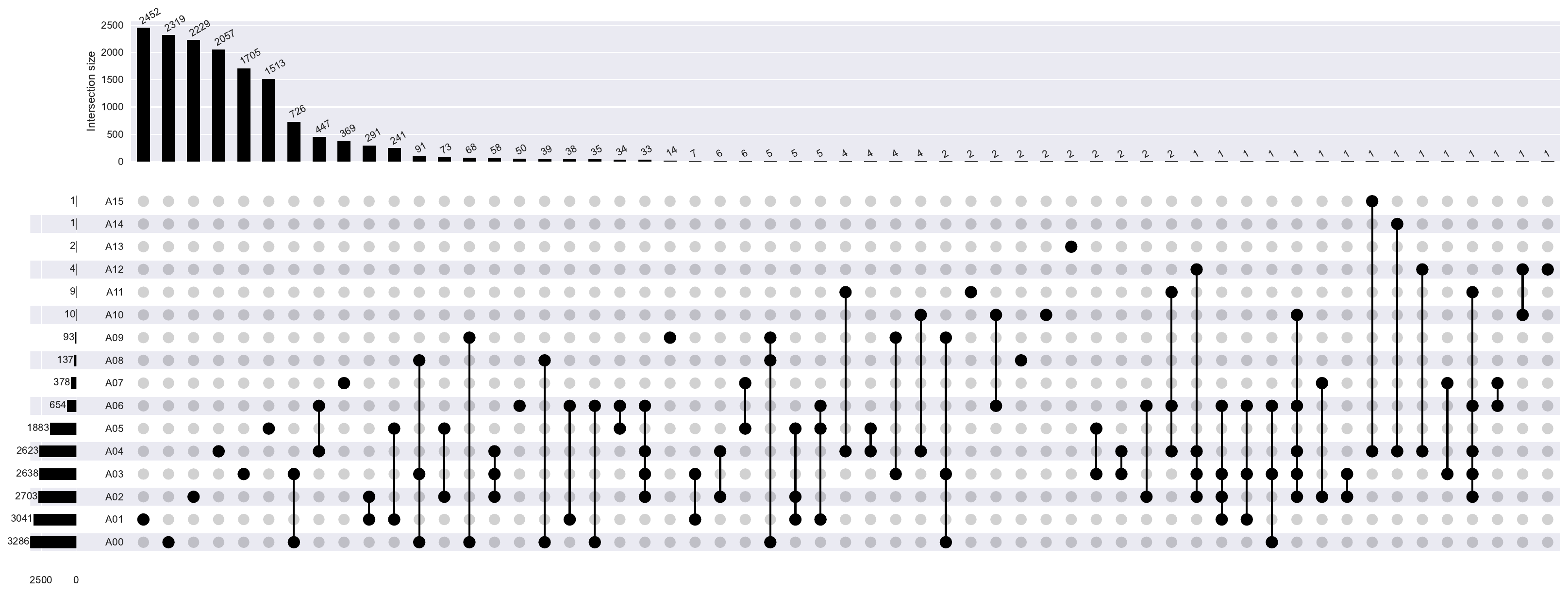}}
\caption{
UpSet plot showing the distribution of segmentations across the \numanntrs \anntrs (\say{A00} -- \say{A15}). The distribution is long-tailed, with the top 6 annotators ($\sim$37\% of the annotators) contributing $\sim$91\% of the segmentations. Best viewed online.
\label{fig:annotator-distribution}
}
\end{figure*}

\begin{figure*}[!htbp]
\centerline{\includegraphics[width=0.9\linewidth]{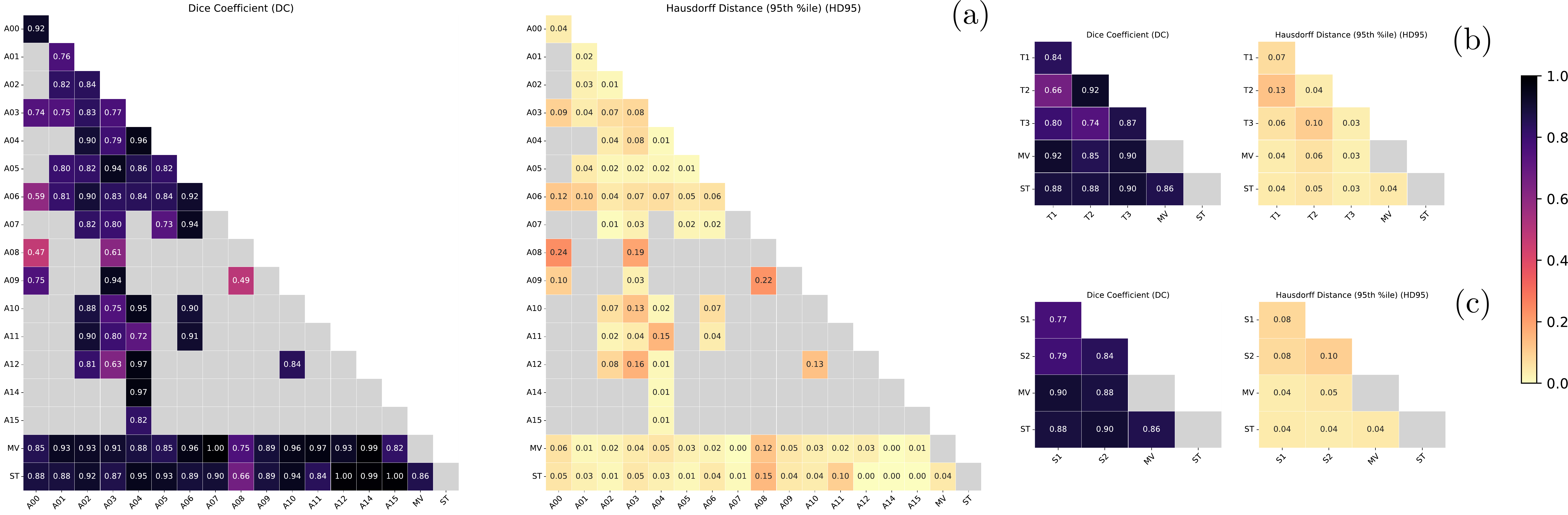}}
\caption{
Quantifying the inter-annotator agreement for \newdatasetname based on the three factors: (a) \anntr, (b) tool, and (c) skill level. For each factor, we report the mean Dice coefficient (left) and 95\textsuperscript{th} percentile of Hausdorff distance (right). Combinations that do not exist in the dataset are grayed out. Best viewed online.
\label{fig:inter-intra-factor-agreement}
}
\end{figure*}

\begin{figure*}[!htbp]
\centerline{\includegraphics[width=0.75\linewidth]{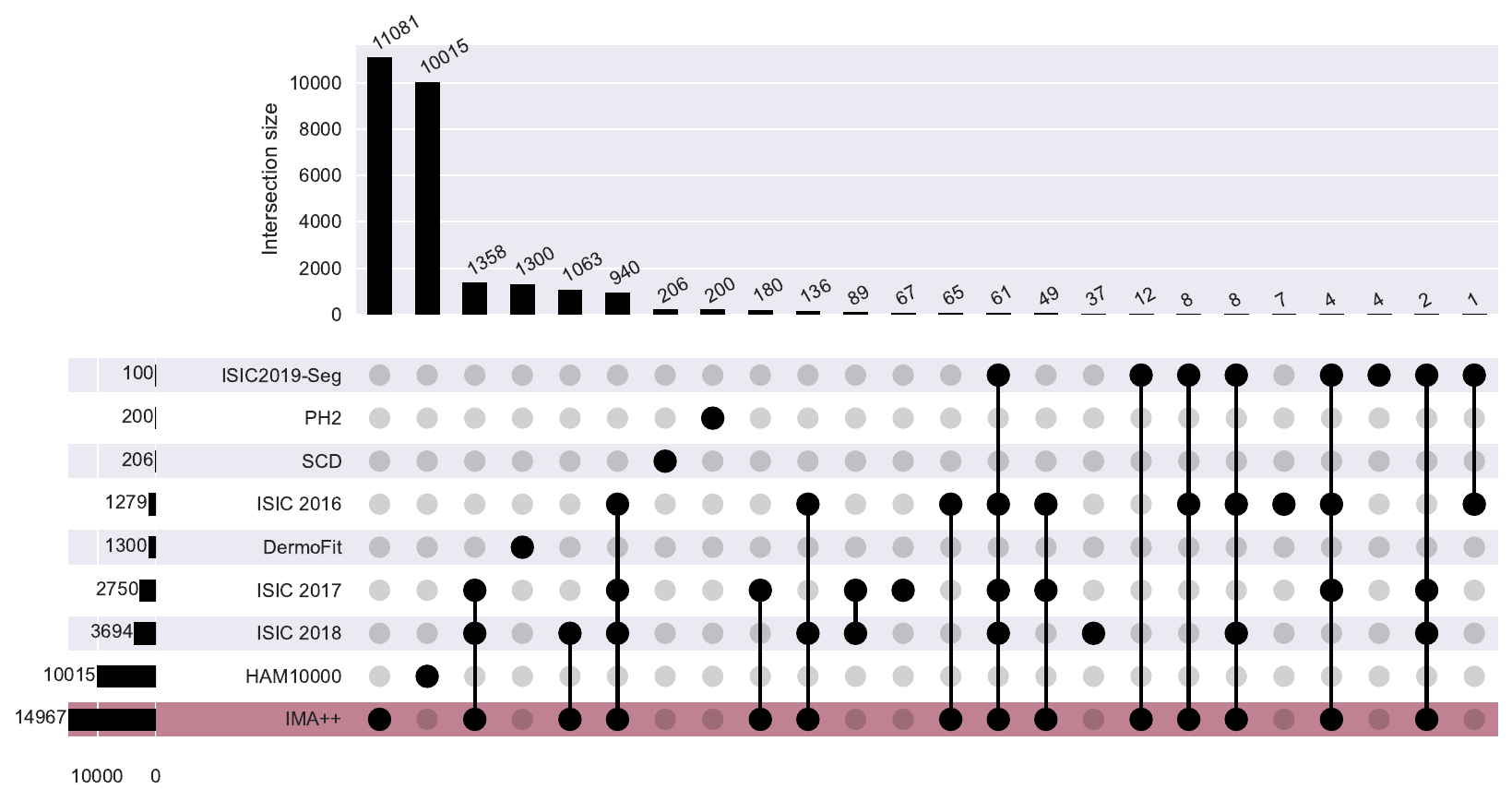}}
\caption{
UpSet plot comparing the proposed \newdatasetname with eight popular public skin lesion image (both dermoscopic and clinical) segmentation datasets and images shared amongst them. With \numimgs, \newdatasetname is the largest dataset, and although it shares some images with other ISIC Challenge Datasets, over 74\% of its images (11,081) are unique from past ISIC challenges. Please also see Table~\ref{tab:datasets-comparison} for more details about these datasets. Best viewed online.
\label{fig:datasets-comparison}
}
\end{figure*}

\section{COLLECTION METHODS AND DESIGN}

\subsection{Data Collection}
The \isicar exposes a public API for automated fetching of its contents. The images and raw \segs in \newdatasetname are originally obtained from the \isicar using the now-deprecated ISIC application programming interface (API) v1~\cite{ISICAPIv1}. The new ISIC API v2~\cite{ISICAPIv2} and its associated command line interface (CLI)~\cite{ISICcli}, however, do not have endpoints for fetching \segs. This leads to a situation where, although the images are available for download on the \isicar, the corresponding \segs or the \seg-associated metadata are not currently available through the API, CLI, or web-interface. To address this gap, we gathered the complete list of \segs and metadata by combining previously downloaded \segs and by contacting sources to obtain the metadata for these masks, followed by anonymization and appropriate organization to build the \newdatasetname dataset.
The scripts used for all the data collection, processing, and analysis are available on GitHub~\cite{imappcode} at \githubrepo.
To facilitate access to the corresponding images, we have curated a dedicated collection on the \isicar~\cite{ISICCollectionIMA} (see Records and Storage).

% . Then we anonymized, organized, analyzed... ... The proposed \newdatasetname dataset and this descriptor address the gap by providing researchers with access to these \segs. 
% The script used to fetch the data is available on GitHub.

\subsection{Data Filtering}
Recent analyses of popular dermatological image datasets have highlighted the importance of rigorous quality assessment, revealing issues such as duplicate images, label noise, and inconsistent metadata that can affect downstream model training and evaluation~\cite{cassidy2022analysis,abhishek2025investigating}.
As an initial quality control for the \seg masks, we subject the masks to the following quality checks: 
% empty masks (\numallblacksegs), masks covering the entire image (\numallwhitesegs), and masks touching the image border (\numsegstouchingborder).
empty masks, masks covering the entire image, masks with disconnected regions, and masks touching the image border.
% \begin{itemize}
%     % \item corrupted mask files: 0,
%     \item empty masks: \numallblacksegs,
%     % \item masks with multiple disconnected regions: 0,
%     \item masks covering the entire image: \numallwhitesegs, and
%     \item masks touching image borders: \numsegstouchingborder.
% \end{itemize}
More specifically, our quality assurance pipeline (implemented in the publicly available \texttt{mask\_qa.py} script~\cite{imappcode}) applies the following checks with associated severity levels: (1) {empty mask detection} (\numallblacksegs masks): masks with zero foreground pixels, indicating a missing or failed \seg; (2) {full-image mask detection} (\numallwhitesegs masks): masks where the entire image is labeled as lesion; (3) {disconnected region detection} (0 masks): masks containing multiple disconnected foreground components, which may indicate annotation errors; (4) {border-touching detection} (\numsegstouchingborder masks): masks where the \seg extends to the image boundary.
% Of these checks, only empty masks \ie masks that do not have any object (lesion) pixels, affect the utility, and therefore we remove these masks from our dataset.
The results of all quality checks are recorded in the \texttt{IMAplusplus\_seg\_metadata\_qa\_results.csv} file included in the GitHub repository~\cite{imappcode}, enabling users to filter masks according to their own quality requirements.
Of these checks, only empty masks \ie masks that do not have any object (lesion) pixels, affect the utility, and therefore we remove these masks from our dataset.

Similarly, we checked for images with missing metadata and found \numimgswithoutmetadata images to not have any associated metadata.
% :
% \begin{itemize}
%     % \item corrupted image files: 0, and
%     \item images with missing metadata: \numimgswithoutmetadata.
% \end{itemize}
For \newdatasetname to be truly useful as a multimodal dataset, all images contained therein should have metadata (Table~\ref{tab:image-metadata}). Therefore, we remove these images, and finally, we are left with \numimgs images that have between 2 and 5 \segs per image, resulting in \numsegstotal \seg masks in total.

Figure~\ref{fig:image-seg-counts} shows the distribution of the number of \segs per image in the \newdatasetname dataset. Of the \numimgs images, \numimgswithsinglesegs have one \seg mask per image, whereas the remaining \numimgswithmultisegs have multiple \segs, leading to a total of $(\numimgswithsinglesegs \times 1) + (2130 \times 2) + (209 \times 3) + (51 \times 4) + (4 \times 5)$=\numsegstotal \segs.

\subsection{Data Processing}

Next, we process the \seg masks and their metadata (Table~\ref{tab:seg-metadata}). First, we assign unique identifiers to the three \anntn \say{factors} that determine the variability in the \seg masks: annotator, tool, and skill level of the manual reviewer.

\noindent\textbf{Annotator mapping:} We sort the \anntrs in decreasing order of the number of \seg masks produced, and assign them unique identifiers accordingly:
% . With 16 annotators in our \newdatasetname, we end up with \anntr IDs: \{\textbf{A00}, \textbf{A01} ..., \textbf{A15}\}.
\{\textbf{A00}, ..., \textbf{A15}\}.

\noindent\textbf{Tool mapping:} \newdatasetname contains three different tools that were used to \say{draw} the \seg masks, and we assign them the following tool IDs:
% \begin{itemize}
%     \item \textbf{T1:} manual polygon tracing by a human expert,
%     \item \textbf{T2:} semi-automated flood-fill with expert-defined parameters, and
%     \item \textbf{T3:} a fully-automated segmentation reviewed and accepted by a human expert.
% \end{itemize}

\begin{itemize}

    \item \textbf{T1:} manual polygon tracing, where the \anntr places a series of polyline control points along the perceived lesion border, and the resulting polygon is converted into a binary \seg mask~\cite{ISICChallenge2016,ISICChallenge2017}.

    \item \textbf{T2:} semi-automated flood-fill, where the \anntr provides a seed point within the lesion and tunes the parameters of a flood-fill algorithm. The algorithm grows a region from the seed point, and is followed by morphological filtering to refine the resulting mask~\cite{ISICChallenge2016,ISICChallenge2017}.

    \item \textbf{T3:} a fully-automated \seg algorithm that generates the initial lesion mask, which is then reviewed and accepted by a human expert. This automated approach was first introduced for the ISIC 2018 Challenge dataset~\cite{ISICChallenge2018}, whereas earlier challenges (2016--2017) used only the manual (T1) and semi-automated (T2) methods~\cite{ISICChallenge2016,ISICChallenge2017}. 
    % The specific algorithm(s) used for T3 are not publicly documented by the \isicar.

\end{itemize}

\noindent\textbf{Skill level mapping:} Next, we have two skill levels of the manual reviewer: \textbf{S1:} expert and \textbf{S2:} novice. These binary labels were part of the \seg metadata provided by the now-deprecated ISIC API v1~\cite{ISICAPIv1} and represent the reviewer's experience.
Third-party tools for downloading \isicar data, such as the ISIC Archive Downloader~\cite{ISICArchiveDownloader} and the \texttt{isicarchive} Python package~\cite{WeberISICArchive} also reference  skill levels only in this binary form.
No finer-grained skill level metadata (\eg years of experience, board certifications) was available through the ISIC API, indicating such granular data was never captured at the source.

The effects of tool choice and skill levels on \seg style are analyzed in Figures~\ref{fig:inter-intra-factor-agreement}b and \ref{fig:inter-intra-factor-agreement}c, respectively.

The \anntr pool is predominantly clinical: 12 \anntrs held a medical degree (MD or MD-PhD), of whom 9 were board-certified dermatologists, while the remaining non-clinical \anntrs all held advanced scientific degrees (MS or PhD).
We report these statistics in aggregate rather than per-\anntr to preserve anonymity, as mapping individual professional details to \anntr IDs could risk re-identification when combined with the published per-\anntr \seg counts (Figure~\ref{fig:image-seg-counts}b). Finer-grained information such as length of experience was not recorded in the \isicar metadata and could not be obtained retrospectively.

Finally, to avoid file corruption issues during data handling, we compute the MD5 hashes of the \seg masks and add them to the metadata, so that users can verify data integrity.

The counts of \segs in \newdatasetname broken down by these three annotation factors is presented in Figure~\ref{fig:image-seg-counts} (b, c, d). 
% Similar to the number of \segs per image, the annotator-\seg counts also exhibit a skewed distribution: six annotators (A00 -- A05) contribute approximately 78\% of the \segs (13,748 out of \numsegstotal). Tools T1 and T3 account for $\sim$81\% of the \segs (14,247 out of \numsegstotal), and $\sim$70\% of the \segs were manually reviewed by an S1 skill level reviewer (12,454 out of \numsegstotal).
Similar to the number of \segs per image, the annotator-\seg counts also exhibit a skewed distribution: six annotators (A00 -- A05) contribute approximately 78\% of the \segs (13,748 of the total \numsegstotal). Tools T1 and T3 account for $\sim$81\% of the \segs (14,247), and $\sim$70\% of the \segs (12,454) were manually reviewed by an S1 skill level reviewer.

\subsection{Computing Segmentation Consensus}
Of the \numimgs images in the dataset, \numimgswithmultisegs images have multiple (2-5) \anntns of the lesion \seg per image. 
To compute a consensus among these \segs, as is standard for multi-\anntr medical image \seg challenges and datasets~\cite{wahid2023large,li2024qubiq,carmo2025manual,wahid2025overview}, we employ two popular consensus algorithms: Simultaneous Truth and Performance Level Estimation (STAPLE)~\cite{warfield2004simultaneous} and majority voting using SimpleITK's \texttt{STAPLEImageFilter} and \texttt{LabelVotingImageFilter}, respectively. 
This allows us to study the agreement between the original \anntns and the consensus masks w.r.t. the \anntn factors (discussed later) and also allows usage of \newdatasetname for the training and evaluation of SLS models without multi-\anntr setups.

Figure~\ref{fig:dataset-samples} shows a few representative samples from the \newdatasetname dataset, with two rows each for images with (from top to bottom) five, four, three, and two \segs per image, respectively. For each of these sets of multi-\anntr masks, the two consensus algorithms' outputs are also shown: majority voting (MV) and STAPLE (ST).

\subsection{Data Splits}

While the availability of consensus masks allows \newdatasetname to be used for lesion diagnosis, single-\seg-per-image SLS, and multi-modal multi-task set ups, the data preparation for multi-\anntr \segs is nuanced and requires careful consideration while splitting the data into training, validation, and testing partitions. Therefore, we split the \numimgswithmultisegs images that have two or more \seg masks per image stratified by 2 criteria:
\begin{itemize}
    \item \textbf{\seg count per image:} The proportions of images with \{2, 3, 4, 5\} \segs per image are similar across the three partitions, yielding (1493, 144, 37, 1) with (2, 3, 4, 5) \anntns in the training partition, with corresponding numbers being (214, 19, 6, 1) and (423, 46, 8, 2) in validation and testing partitions, respectively. 
    \item \textbf{\intanntagr:} We quantify the \intanntagr (IAA) for each image by calculating the per-image averaged pairwise Dice coefficient between \seg masks, and then categorize the images as having low IAA (\ie Dice $\in [0, 0.5)$), medium IAA (\ie Dice $\in [0.5, 0.8]$), or high IAA (\ie Dice $\in (0.8, 1.0]$). We then use these IAA \say{levels} to stratify the splits, ensure that the proportions of these levels are similar across partitions. yielding (165, 403, 1107) images with (low, medium, high) IAA in the training partition,  with corresponding numbers being (24, 58, 158) and (47, 115, 317) in validation and testing partitions, respectively. 
\end{itemize}

Based on these two criteria, we split the \numimgswithmultisegs images into (training, validation, testing) partitions in the ratio of 70:10:20, resulting in (1,675, 240, 479) images across the three partitions (Table~\ref{tab:datasets-comparison}). These standardized data partitions for the multi-\anntr \segs of \newdatasetname allow researchers to systematically report and compare results across methods.

\begin{table}[!htbp]
\centering
\caption{Metadata columns for the images in \newdatasetname.}
\label{tab:image-metadata}
\resizebox{1.0\linewidth}{!}{%
\def\arraystretch{1.25}
\begin{tabular}{@{}cl@{}}
\toprule
Column                                  & Description                                                                                                                                   \\ \midrule
\texttt{isic\_id}                       & \begin{tabular}[l]{@{}l@{}}The unique ID for the image on the ISIC \\ Archive. Example: ISIC\_0010183\end{tabular}                             \\
\hdashline
\texttt{copyright\_license}             & License. Either CC-0 or CC-BY-NC.                                                                                                             \\
\hdashline
\texttt{age\_approx}                    & Approximate age of the patient.                                                                                                               \\
\hdashline
\texttt{anatom\_site\_general}          & \begin{tabular}[l]{@{}l@{}}The general anatomical location of \\ the skin lesion.\end{tabular}                                                \\
\hdashline
\texttt{benign\_malignant}              & The malignancy status of the skin lesion.                                                                                                     \\
\hdashline
\texttt{concomitant\_biopsy}            & \begin{tabular}[l]{@{}l@{}}Whether a biopsy was taken at the \\ same time as imaging.\end{tabular}                                            \\
\hdashline
\texttt{dermoscopic\_type}              & Type of dermoscopic imaging used.                                                                                                             \\
\hdashline
\texttt{diagnosis\_\{1, 2, ..., 5\}} & \begin{tabular}[l]{@{}l@{}}Hierarchical diagnosis labels, \\ wherever applicable.\end{tabular}                                                \\
\hdashline
\texttt{diagnosis\_confirm\_type}         & \begin{tabular}[l]{@{}l@{}}How the diagnosis was confirmed.\\ Example: single image expert \\ consensensus, histopathology, etc.\end{tabular} \\
\hdashline
\texttt{lesion\_id}                      & Unique ID of the skin lesion.                                                                                                                 \\
\hdashline
\texttt{mel\_class}                      & \begin{tabular}[l]{@{}l@{}}Melanoma class, wherever\\ applicable.\end{tabular}                                                                \\
\hdashline
\texttt{mel\_thick\_mm}                   & Melanoma thickness in mm.                                                                                                                     \\
\hdashline
\texttt{melanocytic}                    & Whether the lesion is melanocytic.                                                                                                            \\
\hdashline
\texttt{nevus\_type}                     & Type of nevus, wherever applicable.                                                                                                           \\
\hdashline
\texttt{patient\_id}                       & Unique ID of the patient.                                                                                                                     \\
\hdashline
\texttt{pixels\_\{x, y\}}               & Spatial dimensions of the image.                                                                                                              \\
\hdashline
\texttt{sex}                            & Sex of the patient.                                                                                                                           \\ \bottomrule
\end{tabular}%
}
\end{table}
\begin{table}[!htbp]
\centering
\caption{Metadata columns for the \seg masks in \newdatasetname.}
\label{tab:seg-metadata}
\resizebox{0.8\columnwidth}{!}{%
\def\arraystretch{1.25}
\begin{tabular}{@{}cl@{}}
\toprule
Column                 & Description                                                                                                                                 \\ \midrule
\texttt{ISIC\_id}      & \begin{tabular}[l]{@{}l@{}}The unique ID for the image on the\\ISIC Archive. Example: ISIC\_0010183\end{tabular}                           \\
\hdashline
\texttt{img\_filename} & \begin{tabular}[l]{@{}l@{}}The filename of the image.\\ Example: ISIC\_0010183.JPG.\end{tabular}                                             \\
\hdashline
\texttt{seg\_filename} & \begin{tabular}[l]{@{}l@{}}The filename of the \seg\\ mask. Example:\\ ISIC\_0010183\_A04\_T3\_S2\_\\ 55a9384a9fc3c156bd715c1b.png.\end{tabular} \\
\hdashline
\texttt{annotator}     & \begin{tabular}[l]{@{}l@{}}The ID of the annotator.\\ Example: A04.\end{tabular}                                                            \\
\hdashline
\texttt{tool}          & \begin{tabular}[l]{@{}l@{}}The ID of the tool used.\\ Example: T3.\end{tabular}                                                             \\
\hdashline
\texttt{skill\_level}  & \begin{tabular}[l]{@{}l@{}}The ID of the skill level.\\ Example: S2.\end{tabular}                                                           \\
\hdashline
\texttt{mskObjectID}   & \begin{tabular}[l]{@{}l@{}}The objectID of the \seg\\ mask. Unique for each mask.\\ Example: 55a9384a9fc3c156bd715c1b.\end{tabular}         \\
\hdashline
\texttt{mask\_md5}     & \begin{tabular}[l]{@{}l@{}}The MD5 hash of the \seg\\ mask. Example:\\ f6dae23fab650ba0aa441569a84a7624.\end{tabular}                       \\ \bottomrule
\end{tabular}%
}
\end{table}

\section{VALIDATION AND QUALITY}

\subsection{Visualizing annotator overlap}

We visualize the distribution of the \segs across the \numanntrs \anntrs using an UpSet plot~\cite{lex2014upset} in Figure~\ref{fig:annotator-distribution}. We choose an UpSet plot because relationships among our \numanntrs sets (\numanntrs \anntrs) are too complex to represent with Venn diagrams, which do not scale well beyond 3 sets. Co-occurence matrices are also insufficient, since they only describe pairwise interactions and do not reveal higher-order interactions across multiple sets. In this UpSet plot, the rows correspond to the number of \segs generated by each of the \numanntrs annotators (A00 -- A15). For each row, the cells (denoted by dots) that are part of a set are filled in, and their counts are denoted by the respective bars along the columns on the top. If a column has multiple cells that are filled in, they are connected with a line and the column count denotes the size of the intersection of the corresponding sets. For example, the bottom row corresponds to the annotator A00. Traversing the bottom row shows that there are nine unique sets that A00's \seg masks appear in: 
\begin{itemize}
    \item \{A00\}: 2,319 masks (2\textsuperscript{nd} vertical bar at the top), 
    \item \{A00 $\cap$ A03\}: 726 masks (7\textsuperscript{th} vertical bar), 
    \item \{A00 $\cap$ A03 $\cap$ A08\}: 91 masks, 
    \item \{A00 $\cap$ A09\}: 68 masks, 
    \item \{A00 $\cap$ A08\}: 39 masks, 
    \item \{A00 $\cap$ A06\}: 35 masks, 
    \item \{A00 $\cap$ A08 $\cap$ A09\}: 5 masks, 
    \item \{A00 $\cap$ A03 $\cap$ A09\}: 2 masks, and 
    \item \{A00 $\cap$ A03 $\cap$ A06\}: 1 mask.
\end{itemize}

Since these are the \segs that A00 contributed, their total $2,319 + 726 + 91 + 68 + 39 + 35 + 5 + 2 + 1 = 3,286$ is indicated in the horizontal bar corresponding to A00 (along the bottom row).

With \numanntrs \anntrs, $2^{\numanntrs} - 1$ unique non-empty intersections of annotators are possible. On the other hand, if the images were annotated in a complete bipartite manner, only 1 unique intersection would be possible, \ie all images would be segmented by all the \anntrs. We observe that with an incomplete bipartite annotation set-up, \newdatasetname has 57 unique annotator intersections, yielding a rich variety of multi-\anntr interactions.

\subsection{Analyzing inter- and intra-factor agreement in \newdatasetname}

Studying the extent of (dis)agreement between \anntrs is a commonly explored area with multi-\anntr set-ups. For instance-level labels, this inter-rater agreement (IAA) is often measured using widely used statistics such as Cohen's kappa, Krippendorff's alpha, and Fleiss' kappa. For \seg masks, however, IAA is measured by computing the similarity between the masks, generally using overlap-based (\eg Dice similarity coefficient, Jaccard index) or boundary-based (\eg Hausdorff distance, boundary F1 score) measures. 

\begin{figure}[!ht]
\centerline{\includegraphics[width=0.9\linewidth]{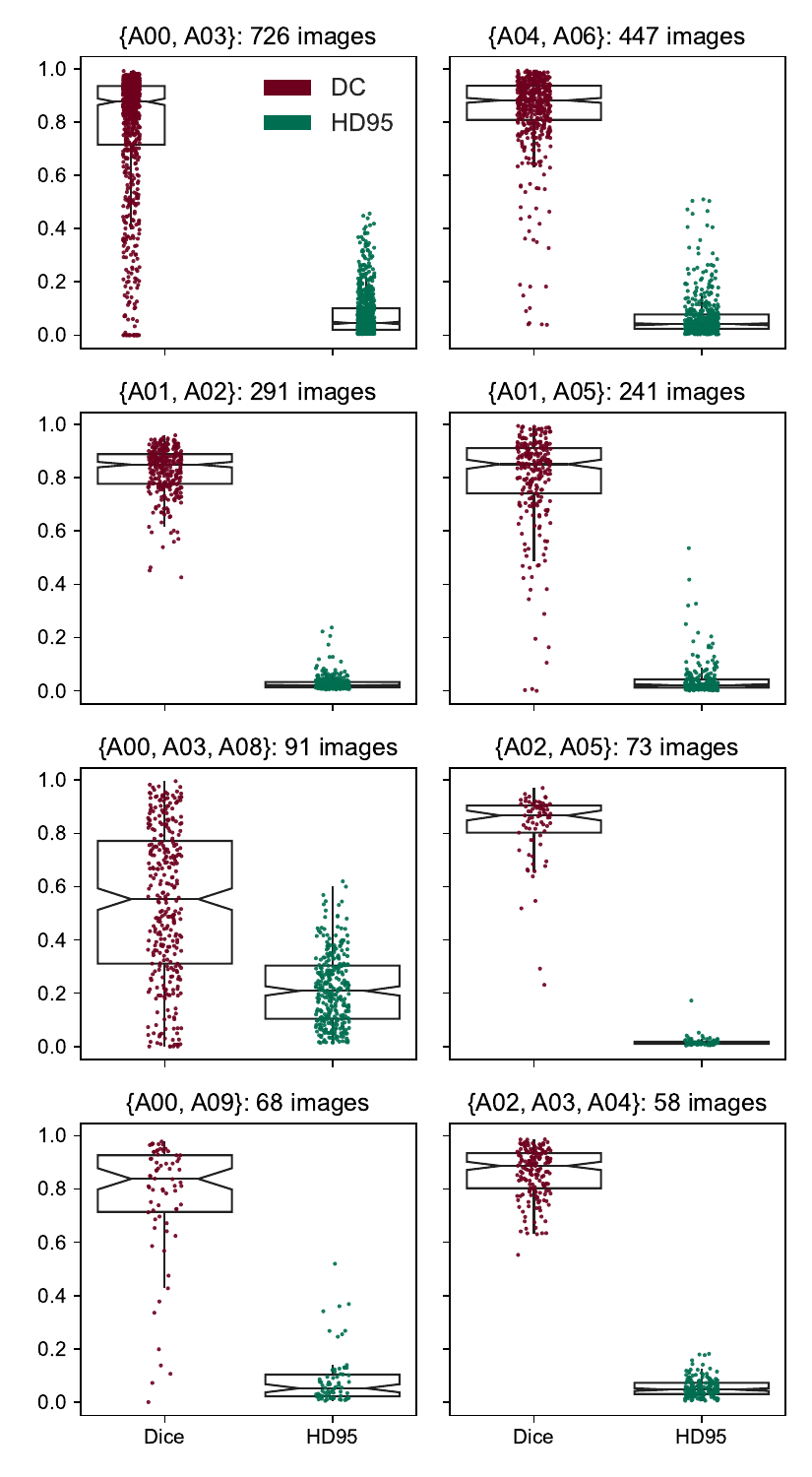}}
\caption{
Inter-annotator agreement distribution, as measured by Dice and HD95, for all combinations of \anntrs that segmented at least 50 images.
\label{fig:annotator-intersection}
}
\end{figure}

\begin{figure}[!htbp]
\centerline{\includegraphics[width=0.9\linewidth]{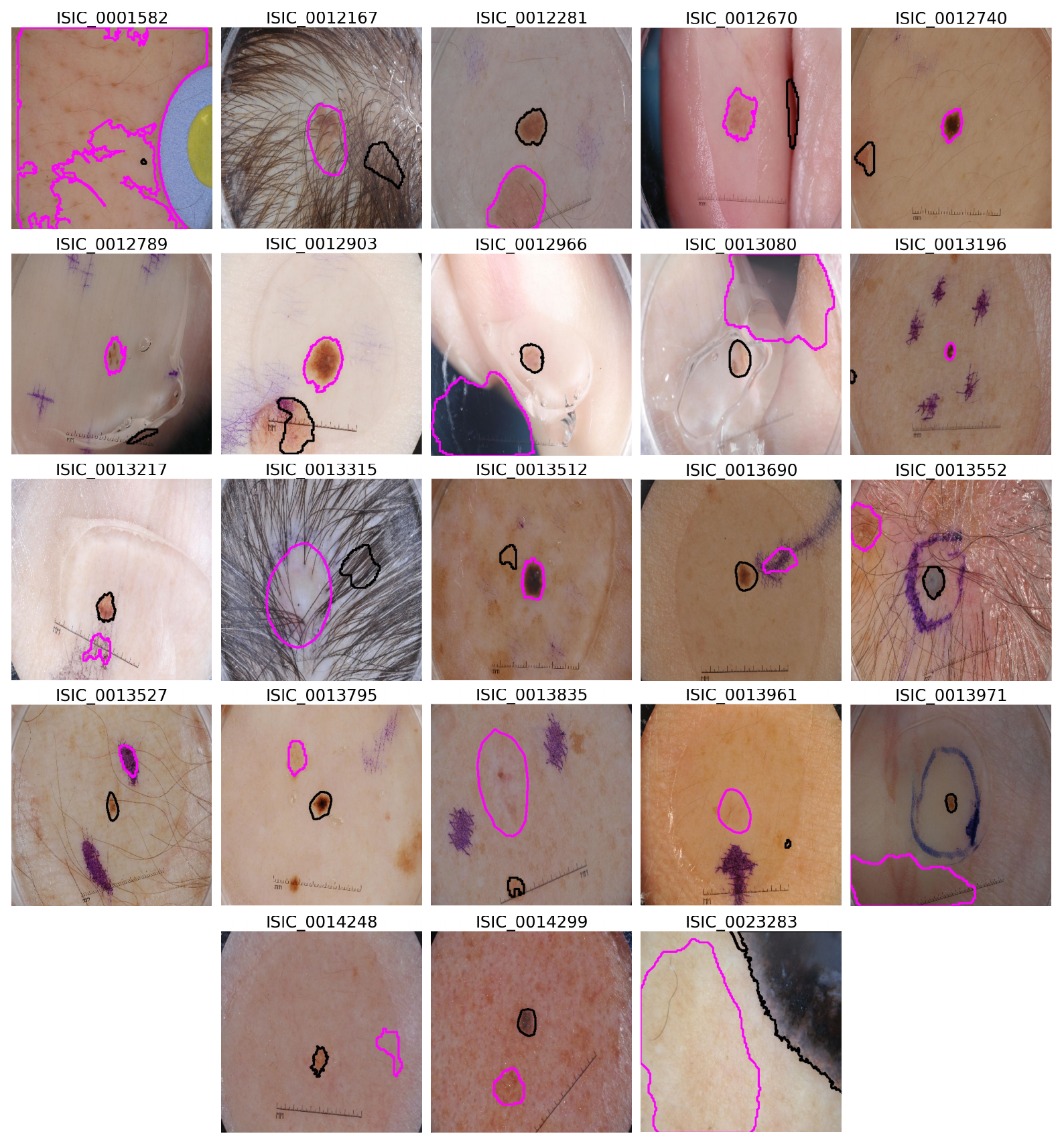}}
\caption{
All images ($n=23$) in \newdatasetname that have entirely non-overlapping \segs (black and magenta contours) from multiple \anntrs. Best viewed online.
\label{fig:zero-overlap}
}
\end{figure}

For our analysis of \newdatasetname, we choose the most popular measure of each of these categories: Dice coefficient (DC) and 95\textsuperscript{th} percentile of the Hausdorff distance (HD95). 
% We extend our analysis beyond just \anntrs, and visualize these measures (DC and HD95) computed between pairs of \segs in Figure~\ref{fig:inter-intra-factor-agreement} for all the factors: \{\anntr, tool, skill level\}, and also include the consensus masks (MV and ST) in our analysis. Note that high DC and low HD95 correspond to higher levels of agreement. Since both DC and HD95 are symmetric, a lower-triangular matrix-based visualization allows us to study both intra-factor (diagonal entries) and inter-factor (off-diagonal entries) agreements. Combinations of factors that are not present in the dataset have been grayed out. There are some key observations from the analysis of \anntrs: (i) some \anntrs (A01, A03) have lower levels of intra-\anntr agreement, (ii) one \anntr (A08) has generally low agreement with all the other \anntrs as well as the consensus masks, (iii) surprising pairs of low (\eg \{A00, A06\}, \{A03, A12\}, \{A04, A11\}) and high (\eg \{A03, A05\}, \{A06, A07\}) agreements emerge, which has also been observed in previous study~\cite{zhao2022skin3d}, and (iv) even majority voting and STAPLE do not exhibit a very high degree of agreement, emphasizing the value of multiple consensus algorithms. A similar pattern emerges when analyzing tools and skill levels. Notably, tools T1 and T3 show much higher agreement with each other than with themselves or T2, whereas T2 exhibits the opposite pattern: high intra-tool agreement but low inter-tool agreement.
Since the images in \newdatasetname vary in the number of pixels, absolute boundary distance metrics (HD, HD95, ASSD) may not be directly comparable across images. Therefore, in addition to the raw metrics, we also compute normalized variants (nHD, nHD95, nASSD) where the raw distance values are divided by the length of the image diagonal, yielding unitless values in $[0, 1]$ that are comparable across images of different sizes. Both raw and normalized metrics are included in the released IAA metric files.
We extend our analysis beyond just \anntrs, and visualize these measures (DC and HD95) computed between pairs of \segs in Figure~\ref{fig:inter-intra-factor-agreement} for all the factors: \{\anntr, tool, skill level\}, and also include the consensus masks (MV and ST) in our analysis. Note that high DC and low HD95 correspond to higher levels of agreement. Since both DC and HD95 are symmetric, a lower-triangular matrix-based visualization allows us to study both intra-factor (diagonal entries) and inter-factor (off-diagonal entries) agreements. Combinations of factors that are not present in the dataset have been grayed out. There are some key observations from the analysis of \anntrs: (i) some \anntrs (A01, A03) have lower levels of intra-\anntr agreement, (ii) one \anntr (A08) has generally low agreement with all the other \anntrs as well as the consensus masks, (iii) surprising pairs of low (\eg \{A00, A06\}, \{A03, A12\}, \{A04, A11\}) and high (\eg \{A03, A05\}, \{A06, A07\}) agreements emerge, which has also been observed in previous study~\cite{zhao2022skin3d}, and (iv) even majority voting and STAPLE do not exhibit a very high degree of agreement, emphasizing the value of multiple consensus algorithms. A similar pattern emerges when analyzing tools and skill levels. Notably, tools T1 and T3 show much higher agreement with each other than with themselves or T2, whereas T2 exhibits the opposite pattern: high intra-tool agreement but low inter-tool agreement.
To provide a quantitative summary of the \intanntagr (A00-A15) across the multi-\anntr subset, Table~\ref{tab:iaa_metrics_summary} reports the mean $\pm$ standard deviation of the pairwise IAA metrics computed over all mask pairs. Additionally, we quantify the agreement between the two consensus algorithms: the mean Dice coefficient between the STAPLE and majority voting consensus masks is $0.8613 \pm 0.1724$, indicating high agreement between the methods. These summary statistics, along with the per-image aggregated metrics, are included in the Zenodo repository as \texttt{iaa\_metrics\_pairwise.csv} and \texttt{iaa\_metrics\_image.csv}, respectively.

\begin{table}[!htbp]
\centering
\caption{{Pairwise inter-\anntr (A00-A15) agreement metrics across the multi-\anntr subset of \newdatasetname (\numimgswithmultisegs images). Boundary distance metrics (HD, HD95, ASSD) are reported in pixels (px), and normalized variants (nHD, nHD95, nASSD) are divided by the image diagonal length. IQR denotes the inter-quartile range.}}
\label{tab:iaa_metrics_summary}
\resizebox{\linewidth}{!}{%
\begin{tabular}{@{}lcc@{}}
\toprule
\textbf{{Metric}} & \textbf{{Mean $\pm$ Std. Dev.}} & \textbf{{Median [IQR]}} \\ \midrule
% % This is summary of metrics for A00-A15 only.
{Dice Coefficient (\textbf{DC})}              & {0.7813 $\pm$ 0.2183} & {0.8572 [0.1879]}        \\
{Jaccard Index (\textbf{JI})}              & {0.6830 $\pm$ 0.2381} & {0.7502 [0.2765]}        \\
{Hausdorff Distance, px (\textbf{HD})}         & {382.5328 $\pm$ 496.5885} & {159.5243 [404.3275]}        \\
{95\textsuperscript{th} percentile Hausdorff Distance, px (\textbf{HD95})}       & {306.2117 $\pm$ 431.7573} & {114.0000 [313.3568]}        \\
{Average Symmetric Surface Distance, px (\textbf{ASSD})}       & {135.8883 $\pm$ 219.2955} & {46.0849 [118.3817]}        \\
{Normalized HD (\textbf{nHD})}             & {0.0970 $\pm$ 0.1046} & {0.0573 [0.0961]}        \\
{Normalized HD95 (\textbf{nHD95})}           & {0.0783 $\pm$ 0.0947} & {0.0408 [0.0769]}        \\
{Normalized ASSD (\textbf{nASSD})}           & {0.0354 $\pm$ 0.0513} & {0.0166 [0.0313]}        \\
\bottomrule

\end{tabular}%
}
\end{table}

\subsection{\newdatasetname versus other datasets}
Table~\ref{tab:datasets-comparison} shows that of these, ISIC 2019-Seg~\cite{zepf2023label} is the only dataset with multiple \anntns per image, however, its small size (100 images, 300 \segs) makes it challenging to both conduct multi-\anntr analysis as well as model the variability among the \segs. On the other hand, while HAM10000's scale (10,015 images with a single \seg per image) is appealing, the lack of multi-expert labels limits its utility beyond traditional single \seg modeling. Among all the single-\anntr and multi-\anntr publicly available SLS datasets covering both dermoscopic and clinical skin imaging modalities, \newdatasetname is the largest with \numimgs images and \numsegstotal total \segs (Table~\ref{tab:datasets-comparison}), allowing researchers to leverage the data for a wide variety of tasks.

Another desirable attribute of \newdatasetname is the number of new skin lesion images for which the \seg masks are unique from past ISIC challenges. The official ISIC Segmentation Challenges' dataset from 2016 through 2018 exhibit a considerable overlap among each other, with 706 images being common across the three datasets~\cite{abhishek2020input}. To conduct a thorough analysis of the overlap among all the public SLS datasets in Table~\ref{tab:datasets-comparison}, we visualize their image identifiers in an UpSet plot~\ref{fig:datasets-comparison}. As expected, we find varying degrees of overlap between ISIC \{2016, 2017, 2018, and 2019-Seg\} datasets. Unsurprisingly, we also discover that a small number of images in \newdatasetname also appear in the previously released ISIC datasets, albeit with only single \seg masks per image for three of these four datasets. However, the majority of the images in \newdatasetname (11,081 images out of total \numimgs; $\sim$74\%) are new, and their \seg masks, single- or multi-\anntr, differ from all the previous ISIC challenge datasets.

\section{RECORDS AND STORAGE} 
\label{sec:records_and_storage}

The list of all files available in the Zenodo repository~\cite{zenodorepo} is presented in Table~\ref{tab:zenodo-filetree}. A single ZIP file contains all the \seg masks in a flat directory structure, which includes all the \numsegstotal masks obtained from the \numanntrs annotators and the consensus masks (majority voting and STAPLE) for all the images with multiple \segs, leading to a total of \numsegstotalwconsensus \seg masks. Since the corresponding skin lesion images are already publicly available and accessible on the \isicar, they are not included in this repository, and can be downloaded using the ISIC API v2~\cite{ISICAPIv2,isicdownloader}.
Moreover, for convenience, we have also created a dedicated collection on the \isicar containing all \numimgs images in \newdatasetname, 
allowing users to download all the corresponding images in a single click~\cite{ISICCollectionIMA}.
The Zenodo repository also includes a comprehensive documentation of the repository contents, as well as direct links to the associated GitHub code repository~\cite{imappcode} and the \isicar collection~\cite{ISICCollectionIMA} for cross-referencing with the processing and analysis scripts.

\begin{table}[!htbp]
\centering
\caption{List of all the files in the Zenodo data repository~\cite{zenodorepo}.}
\label{tab:zenodo-filetree}
\resizebox{\linewidth}{!}{%
\def\arraystretch{1.25}
\begin{tabular}{@{}cl@{}}
\toprule
\textbf{Filename}                                                                                               & \textbf{Description}                                                                                                                                                                          \\ \midrule
\texttt{segs.zip}                                                                                      & \begin{tabular}[l]{@{}l@{}}A ZIP archive of all the \numsegstotalwconsensus \\ \seg masks in \newdatasetname.\end{tabular}                                                                                   \\    \hdashline
\texttt{seg\_metadata.csv}                                                                             & \begin{tabular}[l]{@{}l@{}}A CSV file containing the metadata\\ for all the \numsegstotalwconsensus \seg masks.\end{tabular}                                                                                 \\    \hdashline
\texttt{img\_metadata.csv}                                                                             & \begin{tabular}[l]{@{}l@{}}A CSV file containing the metadata\\ for all the \numimgs skin lesion images.\end{tabular}                                                                         \\    \hdashline
\begin{tabular}[c]{@{}c@{}}\texttt{seg\_metadata\_multi}\\ \texttt{annotator\_subset.csv}\end{tabular} & \begin{tabular}[l]{@{}l@{}}A CSV file containing the metadata\\ for only the multi-annotator subset of \\ \newdatasetname (\ie \numimgswithmultisegs images with\\ multiple \segs per image).\end{tabular} \\    \hdashline
\texttt{iaa\_metrics\_pairwise.csv}                                                                    & \begin{tabular}[l]{@{}l@{}}A CSV file with the IAA metrics\\ calculated for all mask pairs for all\\ the images in the multi-annotator \\subset of \newdatasetname.\end{tabular}                                                        \\    \hdashline
\texttt{iaa\_metrics\_image.csv}                                                                       & \begin{tabular}[l]{@{}l@{}}A CSV file with the pairwise IAA\\ metrics averaged per image.\end{tabular}                                                                               \\    \hdashline
\begin{tabular}[c]{@{}c@{}}\texttt{splits/\{train,val,}\\ \texttt{test\}.csv}\end{tabular}  & \begin{tabular}[l]{@{}l@{}}Standardized \{training, validation,\\ testing\} partitions for the multi-\\annotator subset of \newdatasetname.\end{tabular}                             \\ \bottomrule
\end{tabular}%

}
\end{table}

The \newdatasetname contains rich metadata for both the images and the \segs, whose fields are listed in Tables~\ref{tab:image-metadata} and~\ref{tab:seg-metadata}, respectively. The images' metadata contains patient (\eg age, gender, lesion location) and clinical (\eg type of dermoscopy, how the diagnosis was confirmed, malignancy status, hierarchical diagnosis labels, melanoma thickness, whether a concomitant biopsy was taken) information. The \segs' metadata, on the other hand, contains information previously discussed: unique identifiers about the annotator, tool used, and skill level of the manual reviewer, as well as object identifiers (\texttt{mskObjectID}) and the MD5 hash of the \seg file. The \seg mask files have been richly named to contain all this information, so that even in the absence of a dedicated metadata file, all the necessary information can be fetched directly from the filename. Both these metadata CSV files share a column used to store the unique ISIC identifier of the skin lesion images, and this column can be used as the primary key for merging the two metadata files.
% (JOIN operation) and all other analyses involving both the images and the \segs.

To facilitate data reuse, we report the completeness of the image-level metadata (Table~\ref{tab:image-metadata}): 
99.98\% of images (14,964 images out of total \numimgs) have a recorded diagnosis (any level), 
96.08\% (14,380) have diagnosis confirmation type (\texttt{diagnosis\_confirm\_type}),
99.96\% (14,961) have malignancy status (\texttt{benign\_malignant}),
95.41\% (14,280 images) have sex information, 
94.94\% (14,209 images) have age information (\texttt{age\_approx}), 
and 
32.06\% (4,799 images) have anatomical site information (\texttt{anatom\_site\_general}). 
The diversity of clinical metadata spans multiple anatomical sites, a wide age range, both sexes, and a variety of diagnoses including both melanocytic and non-melanocytic lesions, melanoma and benign conditions, enhancing the generalizability of \newdatasetname for downstream research tasks.
The image and segmentation metadata files are associated via the shared \texttt{ISIC\_id} column, enabling researchers to jointly leverage image-level clinical metadata and segmentation-level annotation metadata for multimodal analyses.

\balance
\section{INSIGHTS AND NOTES}

\subsection{Understanding Zero Overlap Scenarios}

A notable insight from calculating the IAA metrics (DC and HD95) was the wide variability in agreement. Focusing on the multi-\anntr intersections from Figure~\ref{fig:annotator-distribution} with at least 50 images segmented by two or more \anntrs, Figure~\ref{fig:annotator-intersection} shows the distribution of Dice and HD95 for these 8 \anntr sets. We observe a skewed distribution of agreement levels, with a high density at high agreement values and a long tail extending towards low agreements, particularly evident in DC values. This is observed throughout the entire dataset as well: although most images exhibited a reasonably high degree of agreement (DC: 0.866 $\pm$ 0.187; inter-quartile range of 0.161), a substantial number of images showed poor agreement. Specifically, 236 images had mean DC below 0.5, 43 below 0.1, and 23 images had \segs with zero overlap. These 23 images and their corresponding \segs are shown in Figure~\ref{fig:zero-overlap}. Understanding what leads \anntrs to completely disagree on their skin lesions localizations for these 23 images might be a worthwhile direction to explore.

\subsection{Inter-annotator variability, malignancy, and ABCD}

Multiple studies have shown, for example, that lesion \seg variability exists in the clinic, with expert dermatologists favoring \say{tighter} \segs~\cite{silletti2009variability,fortina2012where}. Recently, Abhishek et al.~\cite{abhishek2025what} demonstrated a statistical association between IAA, measured by Dice, and the malignancy of the underlying skin lesion, showing that lower IAA values are linked to malignant lesions. This raises an important question: if clinical prediction criteria such as the ABCD rule rely on an accurate \seg to compute the features, how does variability in \seg impact ABCD calculations and, consequently, the diagnosis derived from these rules? Investigating the effect of \seg variability on clinical decision-making could therefore be an interesting and important direction for future research.

\subsection{Beyond Pairwise Multi-Annotator IAA}

% A limitation of all existing measures for quantifying \intanntagr in medical image \seg is that they are restricted to measuring only pairwise (dis)agreement. Therefore, when reporting image-level agreement values, the ${N \choose 2}$ agreement values computed using pairs of $N$ \segs, any aggregation (\eg mean, median) inevitably leads to the loss of information about the distribution of agreements across the $N$ \segs. A \say{groupwise} IAA measure is missing in the literature and should be developed, and may prove immensely valuable, and \newdatasetname's incomplete bipartite graph of \anntrs serves as an ideal testbed for evaluating the robustness of any such measure.

A limitation of all existing measures for quantifying \intanntagr in medical image \seg is that they are restricted to pairwise (dis)agreement. Consequently, when reporting image-level agreement values, the ${N \choose 2}$ pairwise agreements computed from $N$ \segs must be aggregated (\eg using mean or median), which inevitably loses information about the full distribution of agreements across the $N$ \segs. A truly \say{groupwise} IAA measure is currently missing from the literature, yet its development could be highly valuable. The incomplete bipartite graph of \anntrs in \newdatasetname provides an ideal testbed for evaluating the robustness of such a measure.

\section{SOURCE CODE AND SCRIPTS}

All data collection, processing, validation, and analysis were conducted on an Ubuntu 22.04 workstation with Intel i9-14900K, 64 GB RAM, NVIDIA RTX 4090, with Python 3.10.19. In alphabetical order, the following Python packages with (version numbers) were used: 
\texttt{isic-cli} (12.4.0),
\texttt{matplotlib} (3.10.7),
\texttt{medpy} (0.5.2),
\texttt{numpy} (2.2.6), 
\texttt{opencv-python} (4.12.0), 
\texttt{pandas} (2.3.3), 
\texttt{pillow} (12.0.0), 
\texttt{requests} (2.32.5),
\texttt{scikit-image} (0.25.2),
\texttt{scikit-learn} (1.7.2),
\texttt{scipy} (1.15.3), 
\texttt{seaborn} (0.13.2),
\texttt{simpleitk} (2.5.2),
\texttt{torchvision} (0.24.0),
\texttt{torch} (2.9.0), and
\texttt{upsetplot} (0.9.0).
All scripts used in this work for processing, validation, and analysis are publicly available on GitHub~\cite{imappcode}.
The repository includes an \texttt{overall\_script.sh} that executes the complete pipeline in sequence, and each module has its own \texttt{config.yaml} with all configurable parameters. Step-by-step reproduction instructions and file descriptions are provided.

\section*{ACKNOWLEDGMENTS}
% \hl{Special thanks to Some Company or Some Person for there help on using their API.}
% \\
% \hl{\text{\hspace{1em}} F.A., S.A., and T.A. curated and analysed the data, and wrote parts of the manuscript.
% S.A. reviewed the curation and wrote parts of the manuscript. All authors reviewed the manuscript.}
% \\
% \hl{\text{\hspace{1em}} The article authors have declared no conflicts of interest.}
    % \\
Special thanks to Jochen Weber from Memorial Sloan Kettering Cancer Center for sharing the \segs and the corresponding metadata.

\section*{CONTRIBUTIONS}
K.A. collected, processed, and analyzed the data with inputs and feedback from J.K. and G.H, and wrote the first draft of the manuscript. J.K. and G.H. provided input on the visualization and data presentation, and edited the manuscript. All authors reviewed the manuscript.

\section*{CONFLICTS OF INTEREST}
The authors have no competing interests to declare.
\bibliographystyle{IEEEtran}
% \section{REFERENCES}
% References must follow the standard IEEE format and use numbered citations. Please refer to the IEEE Reference Guide.
\bibliography{references}
% \bibliography{references_short}
\end{document}